\documentclass[11pt]{article}

\usepackage[]{acl}

\usepackage{times}
\usepackage{latexsym}

\usepackage[T1]{fontenc}
\usepackage[utf8]{inputenc}

\usepackage{microtype}

\usepackage{inconsolata}

\usepackage{booktabs}
\usepackage{multirow}
\usepackage{makecell}

\usepackage{graphicx}
\graphicspath{{figures/}}

\usepackage[most]{tcolorbox}
\definecolor{promptframe}{HTML}{8186B8}
\definecolor{promptback}{HTML}{EFF0F8}
\newtcolorbox{promptbox}[2][]{%
  enhanced jigsaw,
  breakable,
  colback=promptback,
  colframe=promptframe,
  colbacktitle=promptframe,
  coltitle=white,
  fonttitle=\bfseries\small,
  title={#2},
  arc=2.5mm,
  boxrule=0.9pt,
  titlerule=0pt,
  left=5pt, right=5pt, top=5pt, bottom=5pt,
  toptitle=2.5pt, bottomtitle=2.5pt,
  before upper={\small\ttfamily\raggedright\setlength{\parskip}{4pt}},
  #1
}

\definecolor{figcc}{HTML}{D99EFD}   % Move 1 - claiming centrality
\definecolor{figmtg}{HTML}{E6C0FE}  % Move 1 - making topic generalizations
\definecolor{figripr}{HTML}{F3E2FE} % Move 1 - reviewing previous research
\definecolor{figig}{HTML}{FD9AC6}   % Move 2 - indicating a gap
\definecolor{figapr}{HTML}{789CF4}  % Move 3 - announcing present research
\definecolor{figop}{HTML}{9FBCFD}   % Move 3 - outlining purposes
\definecolor{figapf}{HTML}{C4D6FE}  % Move 3 - announcing principal findings
\newcommand{\hlstep}[2]{{\setlength{\fboxsep}{1.2pt}\colorbox{#1}{#2}}}

\title{Does AI-Generated Scientific Text Follow Human Argumentation Patterns? A CARS-Based Comparison of Research Article Introductions}

\author{
  \textbf{Abdelrahman Sadallah},
  \textbf{Narjes Sheikh Asadi},
  \textbf{Lonneke van der Plas}
\\
  Università della Svizzera italiana (USI)
}

\begin{document}
\maketitle

\begin{abstract}
  % Abstract text only — wrapped in the abstract environment by acl_latex.tex.
Large language models are moving from helping write up research to helping do
it, which makes it important to know how the scientific text they produce
differs from human writing. Work on this question has stayed mostly at the
surface, using lexical and stylistic cues that light paraphrasing erases. We
look instead at rhetorical structure, the sequence of argumentative moves
through which a text makes its case. We study research-article introductions
under Swales' CARS model, and compare original introductions from published
linguistics articles with generated counterparts of the same papers.
We find that human-written introductions are more flexible in which moves
they use and in what order, while the generated ones are more uniform.
Giving the models the CARS definitions makes them more rigid.

\end{abstract}

% Block top floats in the first column of page 1. Without this, Figure 1 (declared
% at the start of the introduction, i.e. inside this column) floats up to the top of
% the column and lands above the abstract. Blocking the slot sends it to the top of
% the second column. Only affects this column; later columns are unchanged.
\suppressfloats[t]

\section{Introduction}
\label{sec:introduction}

\begin{figure}[t]
  \centering
  \includegraphics[width=\columnwidth]{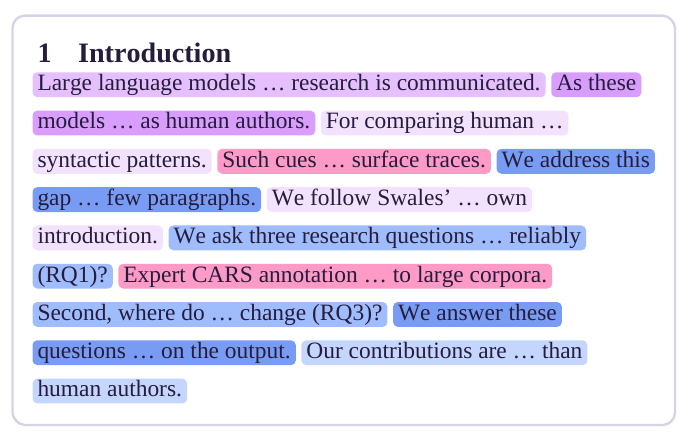}
  \caption{Swales' CARS steps on this paper's own introduction, cut into consecutive \emph{segments},  each
  labeled with one CARS step. A segment's color \emph{hue} represents the parent move, and the \emph{shade} represents the step
  within the move.
  Move~1, establishing a territory (purple):
  \hlstep{figcc}{claiming centrality},
  \hlstep{figmtg}{topic generalizations},
  \hlstep{figripr}{reviewing previous research}.
  Move~2, establishing a niche (magenta):
  \hlstep{figig}{indicating a gap}.
  Move~3, occupying the niche (blue):
  \hlstep{figapr}{announcing present research},
  \hlstep{figop}{outlining purposes},
  \hlstep{figapf}{announcing principal findings}.}
  \label{fig:cars-intro}
\end{figure}

\begin{figure*}[t]
  \centering
  \includegraphics[width=\textwidth]{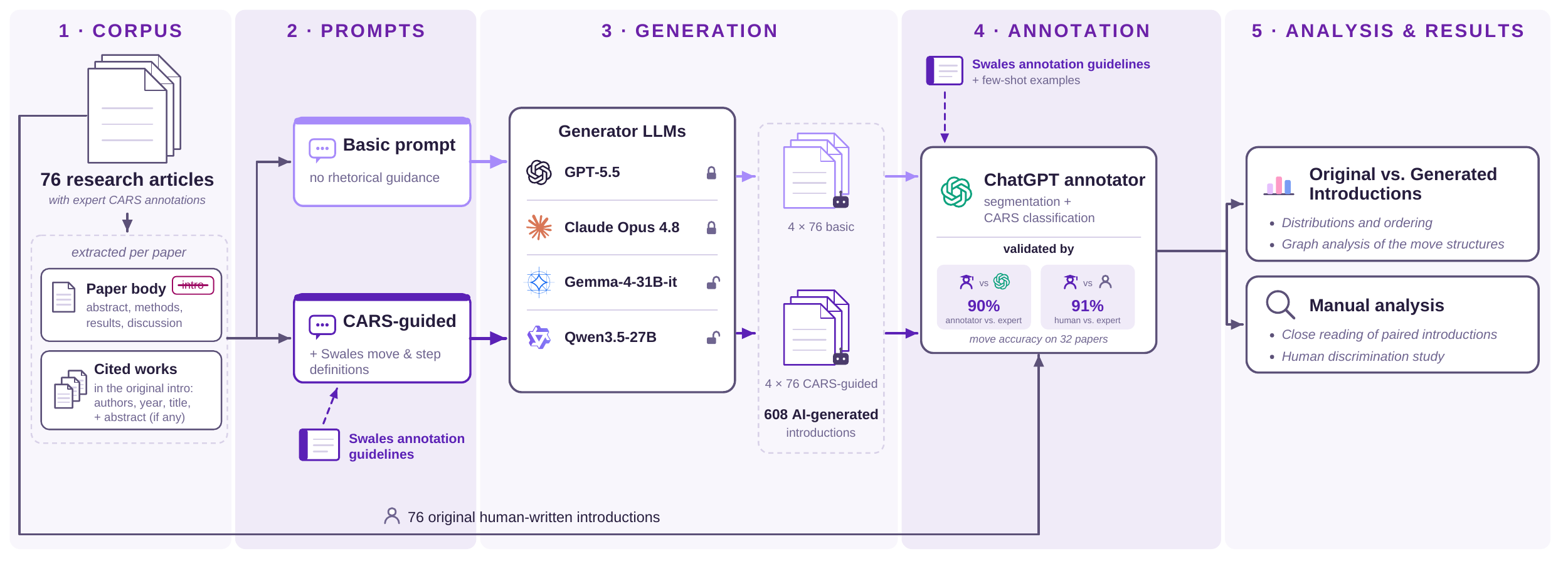}
  \caption{Overview of our framework. From each of the $76$ corpus papers we
  reconstruct the content and the works cited in the original introduction.
  Four LLMs write each paper's introduction under a basic and a CARS-guided
  prompt, giving $608$ generated introductions. An automatic annotator then labels all
  original and generated introductions with CARS moves and steps.
  These labels feed the comparison of the original and generated
  introductions and the manual analysis.}
  \label{fig:framework}
\end{figure*}

Large language models (LLMs) are increasingly involved throughout the
scientific process, from proposing hypotheses and reasoning over the
literature \citep{wang-etal-2023-scientific-discovery,
gottweis-etal-2026-co-scientist} to conducting entire discovery cycles on
their own \citep{lu-etal-2024-ai-scientist} and writing the papers through
which the resulting research is communicated
\citep{van-noorden-perkel-2023-nature, liang-etal-2024-monitoring}. As these
models take on a greater role in scientific writing, it becomes important to
understand how they build an argument and position the work relative to prior
research, and whether they do so in the same way as human authors
\citep{gao-etal-2023-comparing, wu-etal-2025-survey}.

For comparing human and machine-generated text, prior
work relies on lexical statistics, stylistic fingerprints, and syntactic
patterns \citep{gehrmann-etal-2019-gltr,
ippolito-etal-2020-automatic, mitchell-etal-2023-detectgpt}. Such cues say
little about how a model builds a text, and are less suitable for studying scientific writing where what carries the work is the argumentative structure. Moreover, models keep getting better at imitating
human writing style, and light paraphrasing erases most surface traces
\citep{krishna-etal-2023-paraphrasing, sadasivan-etal-2023-canai}.

We address this gap by looking below the surface, at the \emph{rhetorical
structure} of the text: the sequence of argumentative moves that fulfill coherent communicative 
functions within a text~\citep{swales2004research}. We focus on research-article introductions (RAIs),
because they are especially dense in such moves; an introduction has to argue
the paper into existence in the space of a few paragraphs
\citep{swales1990genre, ozturk2007textual}. We follow
Swales' Create-A-Research-Space (CARS) model \citep{swales1990genre,
swales2004research}, a standard account of how introductions are
organised. The CARS model describes an introduction through three rhetorical
moves that typically progress from the general to the specific: the author
establishes a research territory (Move~1), opens a niche in it, for example by
pointing out a gap in previous work (Move~2), and occupies that niche with the
present study (Move~3). Each move is realised through smaller rhetorical units called
\emph{steps}, which realise the move either individually or in
combination~\citep{biber2007discourse},
and in practice moves may recur, and appear in any order, and not
every move need be present~\citep{sheikhasadi2026cars}; Figure~\ref{fig:cars-intro} shows the model at work
on this paper's own introduction.

We ask three research questions. First, can an LLM-based annotator identify the
CARS moves and steps of an introduction reliably (\textbf{RQ1})? Expert CARS annotation
is slow and hard to scale~\citep{biber2007discourse}, so a dependable automatic annotator would open this
kind of rhetorical analysis to large corpora. Second, where do human-written and
LLM-generated introductions differ in their rhetorical structure
(\textbf{RQ2})? And what effect does explicit
CARS guidance have on LLM-generated introductions: when an LLM is given the move
and step definitions, does the rhetorical structure of the RAI it
produces change (\textbf{RQ3})? 

We answer these questions in three parts
(Figure~\ref{fig:framework}). First, we build an automatic annotator that partitions an introduction into
segments by identifying the boundaries of the rhetorical structures at the level of steps and labels each segment with its move and step
(RQ1, \S\ref{sec:method:evaluator}); validated against two human
annotators, ours reaches $92\%$ of the human labelling accuracy and
identifies moves and boundaries at close to human level. Second, we generate introductions at scale
(\S\ref{sec:data:gen}). Starting from $76$ published linguistics articles
whose introductions carry expert CARS annotations \citep{sheikhasadi2026cars}. Four LLMs, two
commercial and two open-weight, write every introduction twice, once
from a basic prompt and once with the CARS move definitions added,
giving $608$ generated introductions. We use the
annotator and human expertise to analyse how the generated introductions differ from the
original ones (RQ2) and what influence CARS guidance has on the output
(RQ3, \S\ref{sec:results}).\footnote{We will release our code
and data upon publication of this work.}

Our contributions are:
\begin{itemize}
  \item We propose to compare human-written and machine-generated
  research-article introductions by their rhetorical structure instead of surface-level features, and show that a human reader tells generated
  introductions from original ones in $90\%$ of blind judgements by relying almost solely on these structures.
  \item We build and validate an LLM-based CARS annotator that segments and
  labels introductions, a reusable instrument for
  researchers who study rhetorical structure at scale.
  \item We present the first rhetorical structure comparison of original and
  LLM-generated RAIs, showing that generated introductions follow a
  narrower and more uniform range of move sequences than human authors.
\end{itemize}

\section{Related Work}
\label{sec:related_work}

Two lines of work bear on our study: the generation and detection of
machine-written scientific text, and the analysis of its rhetorical move
structure.

\subsection{LLMs for scientific writing and its detection}
\label{sec:rw:llm-writing}

LLMs now draft many parts of a paper. They produce abstracts from a title
\citep{gao-etal-2023-comparing}, related-work and citation text
\citep{li-ouyang-2024-related-work, martin-boyle-etal-2024-shallow}, as well as whole papers,
from science-specific models \citep{taylor-etal-2022-galactica} to autonomous
pipelines that ideate, run experiments, and write
\citep{lu-etal-2024-ai-scientist}. A growing fraction of recent papers
contains LLM-modified text \citep{liang-etal-2024-monitoring}.

Detecting machine-generated text usually relies on surface signals: per-token
probability rank \citep{gehrmann-etal-2019-gltr}, probability curvature
\citep{mitchell-etal-2023-detectgpt}, decoding artifacts
\citep{ippolito-etal-2020-automatic}, and watermarking
\citep{kirchenbauer-etal-2023-watermark}.
These signals are fragile. The text that best fools human readers is the hardest
for detectors \citep{ippolito-etal-2020-automatic}. For a strong enough model, the best possible
detector approaches chance \citep{krishna-etal-2023-paraphrasing,
sadasivan-etal-2023-canai}. We therefore look
past the surface to the rhetorical structure of the text, a deeper and possibly more
robust signal that we show humans rely on when determining the source of the introduction.

\subsection{Rhetorical move structure and its analysis}
\label{sec:rw:moves}

Genre analysis treats a RAI as a sequence of conventional
rhetorical moves. Swales' CARS model \citep{swales1990genre, swales2004research}
is the well-established account in this regard: the author first establishes a
territory, then carves out a niche, and finally occupies it. Recovering this move
structure automatically has a long history. Argumentative Zoning assigns each
sentence a rhetorical status \citep{teufel-moens-2002, teufel-etal-2009-towards,
teufel-kan-2011, liakata-etal-2012-zones}. In parallel, work in English for Academic
Purposes (EAP), the applied-linguistics field that teaches academic writing,
has built classifiers for Swales' moves and steps as feedback tools for novice
writers
\citep{anthony-lashkia-2003-mover, pendar-cotos-2008-moves,
cotos-etal-2015-furthering, dayrell-etal-2012-mazea}, and more recent work pairs
expert labels with transformer annotators \citep{raamove2024, omotola-etal-2025-cars}. Closest to us, a recent strand compares
AI-generated and human text by move structure, for the abstract, where
the main gaps are in move frequency and ordering
\citep{kong-liu-2024-comparative, eldakhs-etal-2025}.
For the annotation itself, \citet{kim-lu-2024} prompts ChatGPT to label CARS
moves and steps on applied-linguistics introductions.

We differ from these precedents in both terms of how we make the comparisons and how we annotate the texts. In our comparisons, we  target
the introduction rather than the abstract, as it is the rhetorical and argumentative gateway to a paper~\citep{SheikhAsadiRocci2026ArgumentativePatterns}, compare several models rather than
one, and generate each introduction from the cited literature, an
operationalised approximation of the context a human author writes from.
Studying rhetorical structure at scale also needs labels, and expert CARS
annotation is slow and hard to scale, so a reliable automatic annotator is needed. We extend previous annotation methods
\citet{kim-lu-2024} by benchmarking multiple providers, framing the task as joint
segmentation and classification over the whole introduction instead of
sentence-level labelling, and
validating the annotator against a human--human ceiling.

\section{Materials and methods}
\label{sec:data}

\subsection{Corpus and gold annotations}
\label{sec:data:corpus}
\label{sec:data:gold}

Our starting point is the corpus collected by
\citet{sheikhasadi2026cars} and its manual expert annotations of rhetorical structure: $80$ published research articles from
Applied Linguistics journals, chosen as $20$ papers per journal-prestige quartile
(Q1--Q4). We first verified
that every paper is publicly accessible online; four papers turned out to be
available only behind paywalls, and were discarded. The final corpus comprises the remaining
\textbf{$76$} papers.

We take these annotations, produced against a written annotation manual
\citep{sheikhasadi2025guidelines}, as our \emph{gold} labels for validating
the automatic annotator and answering RQ1. The scheme comprises three moves
and ten steps (Table~\ref{tab:cars}).

\subsection{Generating introductions}
\label{sec:data:gen}

\paragraph{Input context and preprocessing.}
\label{sec:data:preprocessing}
A fair human-vs-AI comparison of \emph{introductions} hinges on what the
model is given. Prior genre comparisons prompt the model with only a short
generic instruction \citep{eldakhs-etal-2025} which is appropriate for an abstract,
that summarises a paper the model can see, but not for an introduction. A
human author writes an introduction from two sources: their own study, and
the knowledge they take from the surrounding scholarly literature they have
read and chosen to cite
\citep{swales1990genre,hyland-1999-attribution,samraj-2005-genre-set}. We
therefore give each model a reconstruction of the information context the
author had: the paper's title and content
and the works cited in the original introduction, each with its
bibliographic metadata and abstract where available. 
To assemble that context from the
published PDFs, we apply a preprocessing pipeline (PDF-to-text conversion,
manual correction, and the identification and enrichment of the works cited
in each introduction), described in
Appendix~\ref{sec:appendix:preprocessing}.
Each generated
introduction is additionally targeted to the length of the paper's original
introduction, so that human--AI differences reflect how the model argues
rather than how much it writes.

\paragraph{Models.}
We generate with four models that cover both commercial and
open-weight systems: two widely used commercial models, GPT-5.5 and Claude
Opus 4.8, and two open-weight models, Gemma-4-31B-it and Qwen3.5-27B, served
locally. The
open-weight condition matters because part of the research community runs
models locally, for instance to avoid sharing unpublished ideas with
commercial APIs. In the rest of the paper we refer to these models by short
names: \textbf{ChatGPT}, \textbf{Claude},
\textbf{Gemma}, and \textbf{Qwen}. Appendix~\ref{sec:appendix:hyperparams}
gives the exact model versions and the settings of every model call.

\paragraph{Generation settings.}

To answer our RQ3, every model writes each
introduction under two settings that differ \emph{only} in rhetorical
guidance. In the \textbf{basic} setting the model is asked to write the
introduction directly, with no guidance about what an introduction should
accomplish rhetorically. In the \textbf{CARS-guided} setting the prompt
additionally supplies the Swales move and step definitions from the
annotation manual and instructs the model to first plan the move sequence
internally (moves may recur, interleave, and appear in any order) and then
write the introduction realising that plan. Both system
prompts are given in Appendix~\ref{sec:appendix:prompts}.

Running the four models under both settings over the $76$ papers yields
$4 \times 2 \times 76 = \textbf{608}$ RAIs.
%\section{Method}
%\label{sec:method}

\subsection{CARS annotation}
\label{sec:method:evaluator}
\paragraph{Automatic annotation.}

Comparing rhetorical structure between human-written and machine-generated
introductions requires CARS annotations on both sides of the comparison, and
human annotation capacity does not scale to this setting: beyond the $76$ human
introductions, our study produces $608$ generated ones (\S\ref{sec:data:gen}),
each of which a trained annotator would have to segment and label. Building
supervised classifiers would in turn require a large labelled training corpus
\citep{raamove2024}. Recent LLMs, however, perform annotation tasks
competitively from instructions and a few demonstrations
\citep{brown-etal-2020-fewshot, gilardi-etal-2023-chatgpt} and have shown
promise specifically for rhetorical move--step analysis \citep{kim-lu-2024}. We
therefore build an LLM-based \emph{automatic annotator} that recovers the gold
annotation format by partitioning an introduction into segments and assigning
each segment one of the ten step labels (\S\ref{sec:data:gold}).

\paragraph{Segmentation and classification.}
Recovering an annotation involves two decisions: \emph{where} the rhetorical
function changes (segmentation) and \emph{which} function each segment realises
(classification). Prompting the model to place boundaries alone systematically
under-segments (mean $6.6$ predicted vs.\ $9.6$ gold segments on a pilot
set~\footnote{set of 16 papers that we did preliminary analysis on.}). Inspired by previous work
on argumentative zoning \citep{teufel-moens-2002}, we also tested classifying
each sentence independently (with its immediate neighbors as context) and
merging adjacent same-label sentences into segments; this removes boundary
placement from the model's job, but it over-segments ($10.5$ predicted
segments/paper vs.\ $9.2$ gold) because it is harder for the model to hold the
same label across the several sentences that belong to one segment.
Asking it to segment and label \emph{jointly} improves boundary
placement (pilot boundary F1 $0.50 \rightarrow 0.71$).  Our final
formulation therefore hands the model the entire introduction and has it emit
an ordered sequence of labeled segments with their classification.

\paragraph{Prompting strategy.}
The system prompt embeds the same annotation manual the human annotators
followed \citep{sheikhasadi2025guidelines}. The full prompt is given in
Appendix~\ref{sec:appendix:prompt}. We also included few-shot demonstrations
\citep{brown-etal-2020-fewshot} which are three fully worked gold annotations chosen
to span the corpus's granularity range: a coarse example ($3$ segments), a
median one ($9$), and a fine-grained one ($18$).
Within the prompt, we add a two-step
decision procedure: identify the parent move first, then the step, and
require a one-sentence rationale before each label, inspired by chain-of-thought prompting
\citep{wei-etal-2022-cot}.
Finally,
we refine the prompt iteratively through error analyses on a held-out set of
$10$ introductions, which are used neither in the few-shot demonstrations nor in
the annotator validation.

In preliminary experiments, we compared ChatGPT, Claude, and Gemini
3.1 Pro. ChatGPT was slightly ahead of Gemini 3.1
Pro and further ahead of Claude on step accuracy ($0.773$ vs.\
$0.762$ and $0.708$). We therefore use ChatGPT
as the annotator for all analyses that follow.

\paragraph{Validating the annotator.}
\label{sec:method:evaluator-validation}

Agreement with the expert gold annotations is hard to interpret on its own: CARS
annotation is a judgement task on which trained humans themselves disagree
\citep{teufel-moens-2002}. We
therefore measure the annotator \emph{relative to a human baseline} obtained
under identical conditions. Two graduate-student annotators with working
knowledge of the CARS framework through the guidelines manual~\citep{sheikhasadi2025guidelines}, independently annotated $32$ introductions
from scratch ($16$ papers each, stratified $8$ per journal quartile). Both the human
annotations and the automatic annotator's output are then scored against the expert gold
(\S\ref{sec:data:gold}) on the same canonical texts.

%\paragraph{Metrics.}
We report two families of metrics one for segmentation and one for classification. Segmentation is scored with boundary F1, WindowDiff, and $P_k$; classification is
scored by the move and step
F1. Appendix~\ref{sec:appendix:validationmetrics} gives the full
definitions.

\paragraph{Annotator performance}
\label{sec:method:evaluator-results}

The annotator reaches $92\%$ of the human label accuracy ($0.694$ vs.\
$0.752$; macro F1 $0.664$ vs.\ $0.699$) and is essentially at the human
baseline on structure: move accuracy $0.901$ vs.\ $0.911$ and boundary F1
$0.725$ vs.\ $0.756$. Table~\ref{tab:evaluator-vs-human} in
Appendix~\ref{sec:appendix:perclass} summarises the validation and gives the
full per-class scores.

Given that the annotator identifies \emph{moves} and places boundaries at
close to human level, labels steps within $6$ points of the human
baseline, and errors mainly where humans also disagree, we adopt it as the
single annotator for all introductions, human-written and generated, in the
analyses that follow (\S\ref{sec:results}. Using one annotator on both sides
removes the annotator as a confound in the human--AI
comparison.
\section{Comparing Original vs.\ Generated Introductions}
\label{sec:results}

\subsection{General distributions and ordering}
\label{sec:results:gen}

\begin{table*}[t]
\centering
\small
\setlength{\tabcolsep}{3.8pt}
\begin{tabular}{lccccccccc}
\toprule
& \textit{Original} & ChatGPT$_b$ & ChatGPT$_c$ & Claude$_b$ & Claude$_c$ & Gemma$_b$ & Gemma$_c$ & Qwen$_b$ & Qwen$_c$ \\
\midrule
\textbf{Move 1: Establishing a Territory} & \textit{59.5} & 70.1* & 62.9 & 57.5 & 51.1* & 60.6 & 46.8* & 54.4 & 48.3* \\
\quad Claiming centrality (CC)  & \textit{18.5} & 21.9* & 20.4 & 19.2 & 17.3 & 19.9 & 17.1 & 16.0 & 15.8 \\
\quad Topic generalization (MTG) & \textit{14.5} & 22.3* & 19.3* & 14.0 & 11.5 & 15.7 &  9.9* & 13.0 & 10.0* \\
\quad Reviewing prev.\ research (RR) & \textit{26.5} & 25.9 & 23.1* & 24.3 & 22.3* & 25.0 & 19.9* & 25.4 & 22.5* \\
\midrule
\textbf{Move 2: Establishing a Niche} & \textit{16.0} & 14.6 & 17.9* & 19.9* & 22.8* & 21.1* & 22.1* & 22.3* & 26.6* \\
\quad Continuing a tradition (CT) & \textit{3.6}  &  3.8 &  4.3 &  3.8 &  4.1 &  6.6* &  6.0* &  4.6 &  4.3 \\
\quad Indicating a gap (Gap)    & \textit{9.6}  &  7.1 &  8.8 & 12.4* & 13.7* & 10.8 & 11.9* & 13.0* & 16.0* \\
\quad Question raising (QR)     & \textit{2.8}  &  3.7 &  4.8* &  3.7 &  4.9* &  3.6 &  4.2 &  4.7* &  6.3* \\
\midrule
\textbf{Move 3: Occupying the Niche} & \textit{24.5} & 15.3* & 19.2* & 22.6 & 26.1 & 18.3* & 31.1* & 23.2 & 25.1 \\
\quad Announcing present res.\ (APR) & \textit{7.1}  &  3.9* &  5.2 &  7.4 &  6.7 &  3.3* &  6.5 &  6.1 &  5.9 \\
\quad Announcing findings (APF) & \textit{3.2}  &  2.3 &  3.9 &  3.1 &  5.2* &  4.1 &  6.7* &  6.0* &  7.3* \\
\quad Indicating structure (IS) & \textit{1.3}  &  0.1 &  0.1* &  0.5 &  1.1 &  0.0* &  6.4* &  0.7 &  0.3 \\
\quad Outlining purpose (OP)    & \textit{13.0} &  9.1* & 10.0* & 11.6 & 13.2 & 10.9 & 11.5 & 10.4 & 11.6 \\
\bottomrule
\end{tabular}
\caption{Distribution of the moves and steps: each cell is the percentage of
all segments produced by the source (pooled over its $76$ introductions) that
carry the given label. \textit{Original} is the human-written introduction.
Subscripts: $b$ = basic, $c$ = CARS-guided prompting.
* marks a value that differs significantly from the original's
(Appendix~\ref{sec:appendix:significance}).}
\label{tab:substeps}
\end{table*}

We first ask
whether the generated introductions reproduce the move and
step \emph{distributions} of the originals, and then whether they
reproduce how the originals \emph{order} and \emph{select} their
steps.
Surface properties of the generated sets, such as their length, number of
move segments, and mean segment length, are reported in
Appendix~\ref{sec:appendix:descstats}. 

\paragraph{Move and step distributions.}
Table~\ref{tab:substeps} reports, for each source, how its move segments
distribute over the three moves and ten steps.

At the move level, the human authors devote $59.5\%$ of their segments to
establishing the territory (Move~1), $16.0\%$ to establishing the niche
(Move~2), and $24.5\%$ to occupying the niche (Move~3). The generated
introductions shift this balance from Move~1 toward Move~2. 
The Move-1 share in turn falls below the human level in five of the
eight sets\footnote{A \emph{set} is the $76$ introductions written by one
model under one prompt, so the four models and two prompts give eight
generated sets; with the $76$ originals this makes the nine sources compared
throughout this section.} (down to $46.8\%$).
This is the most consistent distributional difference we find.

At the step level, the human distribution is heavily skewed. A single
step, \emph{reviewing items of previous research}, accounts for
$26.5\%$ of all human segments, and together with claiming centrality
($18.5\%$), topic generalization ($14.5\%$), and outlining purpose
($13.0\%$) they cover almost three quarters of the human segments. The generated sets keep the same
rank order: reviewing previous research remains the largest step in
every set, and outlining purpose stays within a few points of its human
share ($9.1$--$13.2\%$ vs.\ $13.0\%$).

Finally, comparing each model's basic and CARS-guided versions shows that
the guidance works in the same direction for all four models, as it spends a smaller share of its segments on Move~1
(every Move-1 step shrinks for every model) and a larger share on
Move~2 and Move~3. For three of the four models the guidance
changes the move balance further away from the human one, adding to the
over-use of Move~2 that the generated introductions already show.

\begin{table}[t]
\centering
\small
\setlength{\tabcolsep}{3pt}
\begin{tabular}{lcccc}
\toprule
Source & \makecell{Top step\\3-gram (\%)} & \makecell{Cyclic\\\%}
       & \makecell{Uniq.\\moves} & \makecell{Uniq.\\steps} \\
\midrule
\textit{Original} & \textit{RR${\rightarrow}$MTG${\rightarrow}$RR (25)}
                  & \textit{66} & \textit{2.67} & \textit{5.3} \\
\midrule
ChatGPT$_b$    & CC${\rightarrow}$MTG${\rightarrow}$RR (57) &  70 & 2.93* & 6.3* \\
ChatGPT$_c$    & CC${\rightarrow}$MTG${\rightarrow}$RR (61) &  63 & 3.00* & 6.9* \\
Claude$_b$ & RR${\rightarrow}$CC${\rightarrow}$RR (33)  &  74 & 2.95* & 6.2* \\
Claude$_c$ & CC${\rightarrow}$RR${\rightarrow}$Gap (42) &  51 & 3.00* & 6.6* \\
Gemma$_b$  & RR${\rightarrow}$CC${\rightarrow}$RR (37)  &  67 & 2.96* & 6.1* \\
Gemma$_c$  & CC${\rightarrow}$RR${\rightarrow}$Gap (37) &  50 & 3.00* & 7.4* \\
Qwen$_b$   & RR${\rightarrow}$Gap${\rightarrow}$RR (36) &  95* & 2.97* & 6.8* \\
Qwen$_c$   & CC${\rightarrow}$RR${\rightarrow}$Gap (34) &  87* & 2.99* & 7.0* \\
\bottomrule
\end{tabular}
\caption{Step ordering profile.
\emph{Top step 3-gram}: the source's most frequent sequence of three
consecutive steps, followed in brackets by the percentage of the source's
introductions that contain it.
\emph{Cyclic \%}: percentage of introductions that return to a \emph{move}
they had already left.
\emph{Uniq.\ moves} and \emph{Uniq.\ steps}: mean number of \emph{distinct}
moves (out of $3$) and of \emph{distinct} steps (out of $10$) that one introduction
draws on.}
\label{tab:subseq}
\end{table}

\paragraph{Step ordering and selection}

Table~\ref{tab:subseq} turns from what steps the introductions contain to how
they are ordered. As full step sequences are diverse, we
consider for each introduction its most frequent step 3-gram,
whether it returns to a move it had already left (\emph{Cyclic}),
and how many distinct moves and steps it draws on. One habit is common
to all nine sources: the most frequent opening step is everywhere the
same, claiming centrality. A second commonality is what never occurs: the
scheme's eleventh step, Move-2 \emph{counter-claiming}, is absent from
the human gold annotations and appears in only $4$ of the automatic annotator's
segments, so we exclude it throughout. This highlights that rather than
attacking previous claims head-on, authors and models alike prefer the
softer framing of indicating a gap.

Against the originals, the generated sets trade variety for uniformity.
Humans open with the modal centrality claim in only $46\%$ of the papers,
and another $32\%$ open directly with the literature (RR), an opening the
models almost never choose ($0$--$12\%$); the models start with a
centrality claim in $58$--$84\%$ of their introductions. The humans' most recurrent pattern, a Move-1-internal weave
of reviewing and generalizing (RR$\rightarrow$MTG$\rightarrow$RR), appears
in only a quarter of their papers, whereas each model has a stronger
signature that appears at least 33\% of the time.

Two thirds of the human introductions ($66\%$) return at some point to a
move they had already left, at a rate of $2.4$ returns per introduction.
Half of these returns ($91$ of $181$) go back to Move~1, most often to
review more literature after a first statement of the niche.
The generated sets fall on both sides of that rate.
Basic-prompted ChatGPT, Claude, and Gemma sit at the human rate
($67$--$74\%$), and none of the three differs significantly from the
originals of the same papers. Qwen departs from it upward, cycling in
$95\%$ (basic) and $87\%$ (CARS-guided) of its introductions, significantly
more than the humans.
The CARS-guided sets depart downward, which we return to below(\S\ref{sec:analysis:close}).

What further separates original from generated introductions is the selection: a human introduction
uses on average $2.67$ of the $3$ moves and $5.3$ of the $10$ steps
($28\%$ of the originals omit an entire move), whereas most generated
set realises all three moves and $6.1$--$7.4$ steps. Where a human author
selects from the scheme, the models cover it. Where the original establishes no niche ($19$ papers), states no
explicit gap ($32$), or never claims centrality ($10$), $86$--$96\%$ of
its generated counterparts add the missing step.

Comparing each model's basic and CARS-guided variants shows the guidance
pushing all four models in the same direction, toward more of the scheme
being applied more uniformly, not toward the human profile. Guided
introductions open with a centrality claim even more uniformly. The share of cyclical introductions
falls for all four models (Claude $74 \rightarrow 51\%$, Gemma $67
\rightarrow 50\%$, ChatGPT $70 \rightarrow 63\%$, Qwen $95 \rightarrow 87\%$), and it falls past the human rate of
$66\%$. Guidance also raises the number of distinct steps per introduction for
every model.~\footnote{ This higher count is not greater diversity: each guided
introduction covers more steps, but the guided sets repeat the same
covering, which is why the number of distinct move structures they realise
drops for every model (Table~\ref{tab:graphsim}).}

Finally, in Appendix~\ref{sec:appendix:quartiles} we use the journal-quartile
data to show that the original introductions differ across venues while the
generated sets do not reproduce that variation.

\subsection{Graph analysis of the move structures}
\label{sec:results:graph}

\begin{table}[t]
\centering
\small
\setlength{\tabcolsep}{3.4pt}
\begin{tabular}{lcccccc}
\toprule
& \multicolumn{4}{c}{Similarity to the original} & \multicolumn{2}{c}{Within source} \\
\cmidrule(lr){2-5} \cmidrule(lr){6-7}
Source & Edit$_m$ & Edit$_s$ & Bag$_m$ & Bag$_s$ & Self$_s$ & Str$_m$ \\
\midrule
\textit{Original} & -- & -- & -- & -- & \textit{0.33} & \textit{36} \\
\midrule
ChatGPT$_b$    & 0.62 & 0.41 & 0.83 & 0.63 & 0.43 & 18 \\
ChatGPT$_c$    & 0.61 & 0.41 & 0.81 & 0.62 & 0.44 & 9 \\
Claude$_b$ & 0.61 & 0.42 & 0.81 & 0.63 & 0.40 & 24 \\
Claude$_c$ & 0.61 & 0.43 & 0.80 & 0.61 & 0.46 & 11 \\
Gemma$_b$  & 0.61 & 0.41 & 0.80 & 0.62 & 0.40 & 14 \\
Gemma$_c$  & 0.60 & 0.39 & 0.76 & 0.57 & 0.46 & 8 \\
Qwen$_b$   & 0.55 & 0.38 & 0.78 & 0.60 & 0.37 & 35 \\
Qwen$_c$   & 0.56 & 0.38 & 0.75 & 0.58 & 0.44 & 25 \\
\bottomrule
\end{tabular}
\caption{Graph comparison of move structures
(Appendix~\ref{sec:appendix:graphmetrics} defines the measures). Left:
mean similarity of each generated RAIs to the same paper's
original, as edit similarity (Edit) and bag
similarity (Bag), at the move ($m$) and step
($s$) level. Right: mean pairwise step edit similarity among the source's
own introductions (Self$_s$) and its number of distinct move
structures (Str$_m$).}
\label{tab:graphsim}
\end{table}

The comparisons above rest on pooled distributions;
they say nothing about the similarity between individual introductions and disregard structure.
Here we measure that structural similarity directly, between each generated set and
the originals and within each set. Every introduction is treated as a
labelled chain graph whose nodes are the
CARS classes of the step segments in reading order, each node linked
to the next. Two graphs are compared with two measures, each at the move
($m$) and step ($s$) level: \emph{edit similarity} scores how far the
two class orders agree, and \emph{bag similarity} scores how far the two
introductions give their words to the same classes, with order excluded.
Appendix~\ref{sec:appendix:graphmetrics} defines both measures.

\paragraph{Which model's introductions are most similar to the original
ones?}
Table~\ref{tab:graphsim} (left) reports the mean similarity of each
generated set to the same papers' originals. Higher Bag than Edit similarities show that the models match what an original contains much better than
how it orders it.
What stands out is how low the edit similarities are in
absolute terms: at the step level no generated set exceeds $0.43$, so
even the closest model reproduces well under half of the order in which the
original author moves through the scheme.

\paragraph{How similar are a model's introductions for different papers?}
Table~\ref{tab:graphsim} (right) reports each source's self-similarity,
the mean pairwise step edit similarity among its own $76$
introductions, and the number of distinct move structures it realises. The originals are
the most varied: their self-similarity ($0.33$) is below that of every
generated set ($0.37$--$0.46$), and they realise $36$ distinct move
structures against the models' $8$--$35$. CARS guidance makes every model
less diverse and more self-similar, following the findings of
\S\ref{sec:results:gen}.

\section{Manual analysis}
\label{sec:analysis}

\subsection{Close reading of paired introductions}
\label{sec:analysis:close}

The statistics above say how often the generated structures deviate from the
human ones, but not what a deviation looks like on the page. We therefore
had five papers close-read by a senior graduate student with expertise in
the CARS framework. Each paper
was read as a triplet: the original introduction and
Claude's basic and CARS-guided versions of it, all shown with their
automatic move annotations. The five span all four quartiles and differ in step edit similarity.

\paragraph{Findings}

The generated introductions are more rigidly ordered than the originals, in
two senses.

The first is coverage. All ten generated introductions realize all three
moves, while the originals do not: one of the five never marks a niche at all
and passes from the territory straight to the present study. Where a human
leaves the niche implicit or omits it, the models state it.

The second is order. The generated versions stay closer to the canonical
Move~1--2--3 progression, and where they leave it they leave it in one way,
by returning from Move~2 to Move~1 to review more literature before restating
the niche. Across the corpus, this single transition accounts for $89\%$ of
all returns in the generated sets but only $57\%$ in the originals, which
also go back to Move~1 after announcing the present study. The opening is the clearest case. Human authors
sometimes open with the literature, a topic generalization, or even their
purpose statement, and one of the five originals begins with Move~3. No
generated introduction does, here or anywhere in the corpus: $0$ of the $608$
generated introductions open on Move~3.

One explanation is positional bias, that is, the models place the moves in
the order in which the CARS model presents them rather than in the order the
argument would require. A comparable pull toward the canonical order has been
reported when LLMs are used to analyse moves rather than to write them
\citep{yu-etal-2026-gpt4-moves}. It is consistent with what CARS guidance
does here: the prompt lists the moves in canonical order and says they may
appear in any order, yet the introductions written under it follow the
canonical order more closely than the basic ones (\S\ref{sec:results}), the
opposite of what the guidance was meant to achieve.

% The more interesting finding is that whether this restructuring reads as an
% improvement depends on the original. When the human introduction is thin,
% with one move dominating and a single closing purpose sentence, the generated
% versions appear to be more rich, balancing the three moves and adding steps the
% author never used. When the original is already rhetorically complex,
% realizing all three moves with recursive transitions, the models' appears to be
% straightened into a cleaner funnel: it looks more structured but not more
% effective, and less varied than the human original.

% The models' main effect
% is normalization toward a canonical template, which lifts weak introductions
% and flattens strong ones. CARS guidance amplifies this, using fewer and
% longer segments ($9.9$ against $11.4$ for the basic versions,
% Table~\ref{tab:descstats}).

\paragraph{Authors who know the framework do not write more canonical
introductions.}
A handful of the $76$ articles themselves engage with Swales' CARS model or
rhetorical move analysis, so their authors demonstrably know the framework.
The same expert examined four of them and found that they
follow no common structure. What separates them is the journal quartile
(\S\ref{sec:appendix:quartiles}), not their knowledge of the framework.
Appendix~\ref{sec:appendix:swalesaware} discusses this in more detail.

\subsection{What does a human reader rely on to distinguish between original and generated introductions?}
\label{sec:analysis:detection}

We ask whether the structural differences reported above can be
put to use, that is, whether a reader who is given the CARS model guidelines and a summary of the corpus-level insights can tell an
original introduction from a generated one. The study is blind and was
carried out by a student researcher from outside the project, who judged
$120$ introductions one at a time, $60$ originals and $60$ Claude
generations of the same $60$ papers.
Appendix~\ref{sec:appendix:detection} gives the full protocol.

\paragraph{The judgements rest on structure, not on style.}
The reader classified $108$ of the $120$ ($90\%$) introductions correctly, and
named the argumentation structure as a basis for the decision on $118$ of the
$120$ items against the writing style on $15$ (more than one basis could be
selected per item). They also report having segmented each introduction into moves in their head, and relying on the relations between the moves on top of the moves
themselves, that is, the order and the returns that \S\ref{sec:results:gen}
shows the two sources--human and models--handle differently. They also weighed the corpus-level tendencies they had been given
beforehand about which moves the models over-use and which the human authors
favour. The accuracy therefore comes from the rhetorical structure rather
than from the writing, and the differences highlighted in \S\ref{sec:results} can be used to distinguish between original and generated RAIs.

\paragraph{Introductions written with the CARS-guided prompt are the
easier ones to spot.}
Half of the $60$ generated introductions were produced with the
CARS-guided prompt and half with the basic one. The reader identified
all $30$ CARS-guided ones as machine-written, against $24$ for the
basic. This supports, from the reader's side, what
\S\ref{sec:results:gen} found in the structures themselves: the
CARS-guided prompt makes the introductions more uniform and more rigid
in their ordering, and that uniformity is what makes them recognisable as
machine-written.
\section{Conclusion}
\label{sec:conclusion}

% We compared LLM-written research-article introductions with the human-written
% originals of the same papers by their rhetorical structure, using a validated
% automatic annotator of Swales' CARS moves. Generated introductions are more rigid than the originals, and rigid in
% systematic ways. Where a human author selects from the scheme, the models
% cover it, filling in the moves and steps the original author chose to leave
% out and settling into a narrower range of move sequences than human authors
% use. Supplying the CARS definitions pushes all four
% models further toward the template and away from human structure.
% Because these patterns hold across models and prompts, they are a candidate
% signal for distinguishing generated introductions from human ones. A reader given
% only the CARS framework and a summary of these patterns told the two apart in
% $90\%$ of $120$ blind judgments, and found the CARS-guided introductions the
% easiest to spot. Rhetorical structure therefore carries a usable trace of
% machine authorship, one that sits in the argument rather than in the surface
% style that paraphrasing erases.

We compared LLM-written RAIs with the
human-written originals of the same papers by their rhetorical structure,
using a validated automatic annotator of Swales' CARS moves. The generated
introductions are more rigid, and rigid in the same ways across all four
models used for generation. Where a human author selects some moves, the models cover most of it,
and they settle into a narrower range of move sequences. Supplying the CARS
definitions pushes the models further from human structure. A reader with knowledge of CARS framework and a summary of models' patterns 
could discriminate between original and generated RAIs with 90\% accuracy, with higher performance for RAIs generated using CARS-guided. Rhetorical structure therefore carries a usable trace of
machine authorship, one that sits in the argument rather than in the surface
style that paraphrasing erases.

\section*{Limitations}
\label{sec:limitations}

\paragraph{Fidelity of the input context.}
The information context we give the models (\S\ref{sec:data:gen}) is an
operational approximation of what the author had. First, the cited works are represented by their bibliographic
metadata and abstracts, while the author read the full works. Second, the
citation list is taken from the finished introduction, so the models inherit
the author's selection of the literature; a human author performs that
selection as part of writing.
These choices make the comparison feasible and consistent across papers, but
the models write from a curated subset of the author's knowledge, not from
the same position.

\paragraph{Single discipline and corpus scale.}
Our study covers a single domain, applied linguistics, and a
corpus of $76$ papers released by \citet{sheikhasadi2026cars}. The scarcity of hand-labelled data is in fact what motivated the
automatic annotator we propose (\S\ref{sec:method:evaluator}): rather than annotate
every introduction by hand, we validate an LLM annotator against human labels
and use it to scale the analysis to material that would otherwise be out of
reach. All text is also recovered from PDFs by automatic extraction, which
introduces noise; like \citet{teufel-kan-2011} we operate on extracted
scholarly text, and we mitigate the noise by manual quality control of the parsed
introductions. A larger, multi-domain
corpus would strengthen the statistical claims and let us test how far the
patterns hold beyond this discipline.

\paragraph{Generalisability of the CARS model.}
The CARS model is not a fixed, universal template: the move and step
inventory and the frequency of individual steps vary across disciplines and
even sub-disciplines, which bounds how far our findings transfer beyond applied linguistics. This is visible in our own gold annotations, where
the counter-claiming step is effectively unused. Variability persists within
the discipline as well: \citet{ozturk2007textual} finds that two
applied-linguistics sub-disciplines employ ``different and almost
unrelated'' move structures. Because our corpus pools sub-disciplines, it
may contain heterogeneous move structures, and the results should not be
read as a single canonical ``applied linguistics'' introduction structure.
The more transferable finding is therefore the \emph{gap} between human and
AI move use, not the absolute distributions; the same comparison in another
field could yield different human baselines.

\paragraph{Depth of the argumentation analysis.}
Swales' CARS model captures the rhetorical moves an introduction makes, but
it is not a deep model of argumentation: it labels the function of a passage
(establishing a territory, a niche, or occupying it) without analysing the
internal structure of the reasoning, such as the claims, the evidence, and the
warrants that link them. We treat this work as a first step. Our results
support that reading: even at the level of moves, the framework already
surfaces systematic differences between human and AI introductions, which
shows that Swales' scheme gives usable hints and directions about where the
two diverge. Finer-grained argumentation frameworks, for example the
Argumentum Model of Topics \citep{rigotti2018amt} or a pragma-dialectical
analysis \citep{vaneemeren2018pragmadialectics}, could be layered on top of
this analysis in future work to move from where an argument is made to how it
is constructed, and so yield more detailed findings about the quality of
machine-generated scientific writing.

\paragraph{Annotator and ceiling.}
We use an LLM as the move annotator and validate it against two human
annotators who labelled $32$ introductions between them, $16$ papers each
(\S\ref{sec:method:evaluator-validation}). The human--human ceiling is thus
estimated from a subset of the corpus, and no introduction was labelled by
both annotators. A larger double-annotated set would sharpen the ceiling.

\section*{Ethics Statement}
\label{sec:ethics}

\paragraph{Source material.}
Every paper in our corpus is a published research article that is publicly
accessible online. We checked this for each paper before including it, and we
dropped the four papers whose full text was only available behind a paywall
(\S\ref{sec:data:corpus}). No unpublished, or otherwise private
material was collected or processed. We use the published text only for
analysis, and we do not redistribute the articles themselves.

\paragraph{Annotation work and compensation.}
The human annotations used in this paper (\S\ref{sec:data:gold},
\S\ref{sec:method:evaluator-validation}) come from three kinds of
contributors. Some of the annotators are authors of this paper, and their
annotation work is part of their own research. Others are colleagues who
took part voluntarily as academic collaborators and received no separate
payment for this task. One annotator is a research assistant in our lab,
employed on a salary that covers their working time, of which this
annotation was a part. All annotators knew what the
annotations would be used for and took part with that understanding.

\paragraph{Data handling in the tools we used.}
Two of the four generation models, Gemma and Qwen, were served locally, so
their inputs never left our own machines. The two commercial models were
accessed through the OpenAI and Anthropic APIs, whose terms state that data
sent through the API is not used to train their models. The same holds for
the model we use as an automatic move annotator. Because all of our inputs
are published papers, nothing we sent to an external service was
confidential in the first place.

% \paragraph{What we do not recommend.}
% Everything we sent to an external provider was already published, and our
% setup should not be read as an endorsement of sending draft material to
% these tools. Text that is unpublished, under review, or otherwise
% confidential leaves the author's control once it is sent to a commercial
% service, whatever the stated terms are. Authors who want to work with a
% model on such text should prefer a locally served open-weight model, which
% is part of why we included that condition in our study
% (\S\ref{sec:data:gen}).

\paragraph{Generated introductions.}
We generate introductions to mimic a use case that is already common in
research writing, so that we can study how the resulting argumentation
differs from human writing. This is a research artifact and not a writing
service: none of the generated text is intended for submission anywhere. We
also note that several journals and conferences restrict or forbid
automatically generated content in submitted work.

% Comment out the acknowledgments for the anonymous review version.
% Acknowledgments must NOT appear in the anonymous review version. The
% \input line in acl_latex.tex stays commented out for review; uncomment it
% for the camera-ready version.
\section*{Acknowledgments}

We thank Anastasia Voznyuk for her contribution in annotating
introductions for the validation of the automatic annotator.

\bibliography{custom}

\clearpage
\appendix
\section{The CARS annotation scheme}
\label{sec:appendix:cars}

Table~\ref{tab:cars} lists the three CARS moves and ten steps of the
annotation scheme (\S\ref{sec:data:gold}).

\begin{table}[!ht]
\centering
\small
\begin{tabular}{p{0.30\columnwidth} p{0.58\columnwidth}}
\toprule
\textbf{Move} & \textbf{Steps} \\
\midrule
\textbf{M1}: Establishing a Territory &
Claiming centrality (CC); Topic generalization (MTG); Reviewing previous
research (RR) \\
\addlinespace
\textbf{M2}: Establishing a Niche &
Continuing a tradition (CT); Indicating a gap (Gap); Question raising (QR) \\
\addlinespace
\textbf{M3}: Occupying the Niche &
Announcing present research (APR); Announcing principal findings (APF);
Indicating structure (IS); Outlining purpose (OP) \\
\bottomrule
\end{tabular}
\caption{The Swales CARS move/step taxonomy used for annotation
($3$ moves, $10$ steps). The acronym after each step is the short form
used in the other tables, written as \mbox{move-step} (e.g.\ M1-CC).}
\label{tab:cars}
\end{table}

\section{Preprocessing pipeline}
\label{sec:appendix:preprocessing}

This appendix details the preprocessing pipeline
(\S\ref{sec:data:preprocessing}) that assembles each paper's generation
context from the published PDFs.

\paragraph{PDF-to-text conversion.}
We converted every paper's PDF to structured text with
Docling \citep{auer-etal-2024-docling},\footnote{\url{https://github.com/docling-project/docling}}
selected over pdfminer.six\footnote{\url{https://github.com/pdfminer/pdfminer.six}}
and PyMuPDF\footnote{\url{https://github.com/pymupdf/PyMuPDF}} after manually
inspecting the candidate extractions: Docling best preserved the multi-column
layouts, section headings, and tables.
Structured reference lists, for which layout-oriented extractors perform
poorly, were parsed separately with GROBID
\citep{lopez-2009-grobid}.\footnote{\url{https://github.com/kermitt2/grobid}}

\paragraph{Manual correction.}
We then iterated over the extracted texts and manually fixed the remaining
minor extraction mistakes (broken hyphenation, mis-ordered text blocks, and
running headers or footers leaking into the body), yielding a clean full-text
version of all $76$ papers.

\paragraph{Introduction citations and their abstracts.}
We next identified every work cited in each \emph{introduction}: in-text
citations were matched to reference-list entries automatically using
author-year citation patterns, and the result was then manually verified and
corrected paper by paper. Across the
corpus this yields roughly $1{,}065$ cited works (a mean of $14$ per
introduction). Finally, we retrieved bibliographic metadata and abstracts for
these works from multiple scholarly databases:
OpenAlex,\footnote{\url{https://openalex.org}}
Crossref,\footnote{\url{https://www.crossref.org}} and Semantic
Scholar,\footnote{\url{https://www.semanticscholar.org}} complemented by a
cascade of further academic APIs for hard-to-find entries. Abstracts were
recovered for about $72\%$ of the cited works; for the remainder, we enter the
generation context with their title and authors only.

\section{Model configuration}
\label{sec:appendix:hyperparams}

The introductions (\S\ref{sec:data:gen}) were generated with
\texttt{gpt-5.5}, called through the OpenAI Batch API on 7 July 2026,
\texttt{claude-opus-4-8}, called through the Anthropic Message Batches API on
6 July 2026, and \texttt{google/gemma-4-31B-it} and
\texttt{Qwen/Qwen3.5-27B}, served locally with vLLM
\citep{vllm}.\footnote{\url{https://github.com/vllm-project/vllm}} The
automatic annotator (\S\ref{sec:method:evaluator}) is \texttt{gpt-5.5}, also
through the Batch API, and the preliminary annotator comparison additionally
used \texttt{gemini-3.1-pro-preview}. Every call uses the provider's default
decoding settings and low reasoning effort.

\section{Annotator validation results}
\label{sec:appendix:perclass}

\begin{table}[t]
\centering
\small
\setlength{\tabcolsep}{4pt}
\begin{tabular}{lccc@{\hspace{12pt}}c}
\toprule
& \multicolumn{3}{c}{\textit{Classification}} & \textit{Segmentation} \\
\cmidrule(lr){2-4} \cmidrule(lr){5-5}
Annotator & step-acc & step-F1 & move-acc & bound-F1 \\
\midrule
HB & 0.752 & 0.699 & 0.911 & 0.756 \\
AA & 0.694 & 0.664 & 0.901 & 0.725 \\
\bottomrule
\end{tabular}
\caption{The automatic annotator (AA) vs.\ the human baseline (HB), both
scored against the expert gold on the same $32$ introductions. The two
column groups score the two decisions separately.
\textit{Classification} scores \emph{which} rhetorical function a piece of
text realises: step-acc and step-F1 are the accuracy and the
macro-averaged F1 of the per-sentence step label, and move-acc is the same
accuracy after aggregating each sentence's step to its parent move.
\textit{Segmentation} scores \emph{where} the rhetorical function changes,
independently of the labels: bound-F1 is boundary F1 with a $\pm5$-character
tolerance. Appendix~\ref{sec:appendix:validationmetrics} gives the full
definitions.}
\label{tab:evaluator-vs-human}
\end{table}

Table~\ref{tab:evaluator-vs-human} reports the validation of the automatic
annotator against the human baseline
(\S\ref{sec:method:evaluator-validation}).

Table~\ref{tab:perclass} gives per-class F1 for the ten CARS steps under
the per-sentence label metric, for the human second annotators and the
automatic annotator, both scored against the expert gold on the $32$
validation introductions (\S\ref{sec:method:evaluator-validation}). The
human--annotator gap is concentrated in claiming centrality (M1-CC) and, to a
lesser extent, making topic generalization (M1-MTG) and announcing principal
findings (M3-APF); the two sides are at parity on the two most frequent
classes (M1-RR, M3-OP) and on indicating a gap (M2-Gap).
\begin{table}[t]
\centering
\small
\begin{tabular}{lcc}
\toprule
Class & Human & Annotator \\
\midrule
M1-CC  & 0.70 & 0.46 \\
M1-MTG & 0.63 & 0.55 \\
M1-RR  & 0.86 & 0.84 \\
M2-CT  & 0.23 & 0.32 \\
M2-Gap & 0.77 & 0.77 \\
M2-QR  & 0.42 & 0.35 \\
M3-APR & 0.62 & 0.63 \\
M3-APF & 0.77 & 0.62 \\
M3-IS  & 0.92 & 0.92 \\
M3-OP  & 0.85 & 0.81 \\
\bottomrule
\end{tabular}
\caption{Per-class F1 on the $32$-paper validation set,
scored against the expert gold. \textit{Human} is the second-annotator
baseline; \textit{Annotator} is our automatic annotator. Classes are
named \mbox{move-step} as in Table~\ref{tab:cars}.}
\label{tab:perclass}
\end{table}

\section{Evaluation metrics}
\label{sec:appendix:metrics}

\subsection{Annotator validation metrics}
\label{sec:appendix:validationmetrics}

These are the metrics used to validate the automatic annotator against the human
baseline (\S\ref{sec:method:evaluator-validation}). They form two families that
decouple \emph{where} the moves change from \emph{which} move each segment
realises.

\paragraph{Segmentation (label-independent).}
\textbf{Boundary F1} matches predicted to gold boundaries within a
$\pm5$-character tolerance. We also report \textbf{WindowDiff}
\citep{pevzner-hearst-2002-critique} and \textbf{$P_k$}
\citep{beeferman-etal-1999-statistical}, the standard text-segmentation error
scores that credit near-miss boundaries gracefully.

\paragraph{Classification.}
We compute classification metrics at the sentence level rather than the segment
level, because the predicted and gold segments can differ in number; this is
also standard practice in the literature on argumentative zoning and CARS
move/step classification \citep{teufel-moens-2002, teufel-kan-2011,
pendar-cotos-2008-moves, cotos-etal-2015-furthering, raamove2024}. Each sentence
in the introduction is assigned the predicted and gold labels of the segment it
belongs to, and we report sentence-level label accuracy. Finally, we aggregate
each sentence's step to its parent move and report \emph{move accuracy}.

\subsection{Move-structure similarity measures}
\label{sec:appendix:graphmetrics}

We define the two measures used in the graph analysis
(\S\ref{sec:results:graph}). Both compare two introductions through the
chain graphs of their move segments, and both are computed at the move
level ($3$ classes) and the step level ($10$ classes).

\paragraph{Edit similarity.}
Edit similarity scores how far two introductions order their classes the
same way:
\[
\mathrm{EditSim}(a,b) \;=\; 1-\frac{\mathrm{Lev}(a,b)}{\max(|a|,|b|)},
\]
with $\mathrm{Lev}$ the Levenshtein distance
\citep{levenshtein-1966-binary} between the class sequences $a$ and $b$:
the minimum number of insertions, deletions, and substitutions that turns
one sequence into the other, normalised by the longer sequence's length so
the score lies in $[0,1]$. Because the graphs are chains, the Levenshtein
distance between the label sequences is exactly the graph edit distance
between the graphs \citep{sanfeliu-fu-1983-distance}. Before the
comparison, consecutive segments carrying the same class are merged into
one node (the analysis convention of \S\ref{sec:appendix:descstats}).

\paragraph{Bag similarity.}
Bag similarity ignores the order and asks only how much of an introduction
goes to each class. Every introduction divides its words among the classes,
which gives one share per class; the shares add up to one and form a
distribution $p$. Two introductions are then compared by how much of these
shares they spend the same way:
\[
\mathrm{BagSim}(p,q) \;=\; \sum_{c} \min(p_c, q_c),
\]
that is, for each class we keep the smaller of the two shares and add these
up, which is the part of the two profiles that overlaps. If one introduction
gives $60\%$ of its words to Move~1 and the other gives $40\%$, that class
contributes $0.4$ to the sum. The score is $1$ when the two introductions
divide their words over the classes in exactly the same way and $0$ when
they have no class in common. This is the histogram intersection of
\citet{swain-ballard-1991-color}, known in statistics as the overlapping
coefficient \citep{weitzman-1970-overlap, inman-bradley-1989-overlapping};
for two distributions it equals one minus the total variation distance
between $p$ and $q$.

A class's share is its share of the introduction's \emph{words}, not its
share of the segments. The reason is that the same text can be cut into more
or fewer segments: an annotator that splits one long stretch of literature
review into three segments would make that class look three times as
important if we counted segments, while its word share does not move. Read
together,
the two measures separate content from arrangement: a pair with high bag
similarity but low edit similarity gives its words to the same rhetorical
classes at similar weights but orders them differently.

\section{Significance testing}
\label{sec:appendix:significance}

All comparisons between an original and a generated set are paired by paper:
the two introductions compared are always of the same paper, so the paper is
the unit of analysis ($n=76$). We test them with the Wilcoxon signed-rank
test \citep{wilcoxon-1945-individual}, except for cyclicity, which is a
per-introduction binary and is tested with the McNemar test
\citep{mcnemar-1947-note}. A star marks $q<0.05$. 
% The
% tables report shares pooled over all segments a source produces, whereas the
% tests operate on per-introduction shares; the two agree in direction for
% every starred cell.

The model comparison of \S\ref{sec:results:graph} is tested the same way, on
each paper's step edit similarity to its own original. 

\section{Descriptive statistics}
\label{sec:appendix:descstats}

Table~\ref{tab:descstats} compares the original human-written introductions
with the eight generated sets on three surface properties: length, number of
segments, and mean segment length, with every set annotated by the
automatic annotator (\S\ref{sec:method:evaluator}).

For the generation length, ChatGPT overshoots the
target by roughly $20\%$ while the other models stay close. The clearest
structural difference is in rhetorical granularity. Throughout the
analyses, consecutive segments that carry the same step label are
counted as one segment, so a new segment always marks a change of
rhetorical function. Every generated set realises more segments than
the human originals ($9.9$--$13.9$ segments per introduction vs.\ the
human $9.4$), meaning the models switch rhetorical function more often
than human authors. All sets except CARS-guided Claude also use shorter
segments ($50.2$--$57.3$ words per segment vs.\ the human $65.1$), while
CARS-guided Claude develops its segments at human length ($66.1$). CARS
guidance moves Claude and Qwen visibly closer to the human granularity
(e.g.\ Claude $11.4 \rightarrow 9.9$ segments), but leaves ChatGPT and
Gemma essentially unchanged.

\begin{table}[t]
\centering
\small
\setlength{\tabcolsep}{4pt}
\begin{tabular}{lccc}
\toprule
Source & \makecell{Words\\/intro} & \makecell{Segments\\/intro} & \makecell{Words\\/segment} \\
\midrule
\textit{Original (human)} & \textit{615} & \textit{9.4} & \textit{65.1} \\
\midrule
ChatGPT, basic & 743 & 13.9 & 53.5 \\
ChatGPT, CARS  & 745 & 13.7 & 54.4 \\
Claude, basic  & 655 & 11.4 & 57.3 \\
Claude, CARS   & 653 &  9.9 & 66.1 \\
Gemma, basic   & 563 & 10.5 & 53.7 \\
Gemma, CARS    & 590 & 10.5 & 56.0 \\
Qwen, basic    & 659 & 13.1 & 50.2 \\
Qwen, CARS     & 612 & 11.7 & 52.4 \\
\bottomrule
\end{tabular}
\caption{Descriptive statistics of the original and generated introductions
($76$ each): mean words per introduction, mean number of segments, and
mean words per segment.}
\label{tab:descstats}
\end{table}

\section{Variation across journal quartiles}
\label{sec:appendix:quartiles}

The corpus is stratified by journal quartile (Q1--Q4,
\S\ref{sec:data:corpus}), which lets us ask whether the rhetorical profile
of the originals varies with publication venue, and whether the generated
introductions follow that variation. Table~\ref{tab:quartilerep} reports,
per quartile, how much of the CARS repertoire an introduction uses (the
mean number of distinct moves and distinct steps) and how often it
cycles (the percentage of introductions that return to an already-visited
move). The original introductions differ across quartiles: Q1 papers use
almost all three moves ($2.88$ on average) and cycle the most ($75\%$),
while Q3 papers use the fewest moves ($2.50$) and cycle the least
($55\%$); Q2 papers draw on the widest range of steps ($5.7$ of the
$10$).

The generated introductions show much less variation. In every quartile
they use nearly all three moves ($2.97$--$2.99$), more steps than the
originals ($6.1$--$7.2$ vs.\ the human $5.0$--$5.7$), and cycle at a
nearly constant rate ($66$--$74\%$, against the human range of
$55$--$75\%$). The quartile differences visible in the originals are
therefore not reproduced. This is expected: the models write in a
uniform style, and we gave them no information about the journal quartile of
the paper they were writing for.

\begin{table}[t]
\centering
\small
\setlength{\tabcolsep}{4.5pt}
\begin{tabular}{llccc}
\toprule
 & Source & \makecell{Uniq.\\moves} & \makecell{Uniq.\\steps}
        & \makecell{Cyclic\\\%} \\
\midrule
\multirow{2}{*}{Q1} & Original  & 2.88 & 5.0 & 75 \\
                    & Generated & 2.99 & 6.1 & 73 \\
\midrule
\multirow{2}{*}{Q2} & Original  & 2.65 & 5.7 & 70 \\
                    & Generated & 2.97 & 7.2 & 74 \\
\midrule
\multirow{2}{*}{Q3} & Original  & 2.50 & 5.3 & 55 \\
                    & Generated & 2.97 & 6.7 & 66 \\
\midrule
\multirow{2}{*}{Q4} & Original  & 2.70 & 5.3 & 65 \\
                    & Generated & 2.98 & 6.5 & 66 \\
\bottomrule
\end{tabular}
\caption{CARS repertoire and cyclicity by journal quartile.
\emph{Uniq.\ moves}: mean number of \emph{distinct} moves (of $3$) per
introduction; \emph{Uniq.\ steps}: mean number of \emph{distinct} steps
(of $10$) per introduction; \emph{Cyclic \%}: percentage of introductions
that return to a \emph{move} they had already left. \emph{Generated}
pools the eight generated sets.}
\label{tab:quartilerep}
\end{table}

\section{The Swales-awareness effect}
\label{sec:appendix:swalesaware}

A handful of the $76$ articles themselves engage with Swales' CARS model or
rhetorical move analysis, so their authors demonstrably know the framework.
Does that knowledge show in how they build their own introductions? The
expert of \S\ref{sec:analysis:close} examined four such articles, one from
each journal quartile. This appendix gives the readings summarised there.

Knowing the framework does not make these authors write more canonical
introductions: none of the four follows the standard Move~1--2--3 funnel.
Two never mark an explicit niche and pass from the territory straight to
their purpose statement. A third does raise a niche but folds it into a
territory statement instead of giving it its own move, so a niche that is
rhetorically present is not counted as Move~2. A fourth returns to the
territory after announcing its work and ends on Move~1. Their command of
the model is real, but it does not surface as a more schematic structure.

What does track structure is the journal quartile, an association also
reported across high- and low-impact venues
\citep{alanazi-alqarni-2022-first-sentence}. In our corpus
(\S\ref{sec:appendix:quartiles}), higher-quartile introductions move among
the three moves more freely and realize Move~3 more fully, whereas
lower-quartile ones lean on Move~1 and compress or omit the later moves.
The quartile is a stronger predictor of an introduction's structure than
the author's knowledge of the CARS model.

\section{Human detection study}
\label{sec:appendix:detection}

This appendix details the detection study of \S\ref{sec:analysis:detection},
in which an informed but blind reader judges, one introduction at a time,
whether it was written by a human or generated by a model.

\paragraph{Reader.}
The judge is a student researcher who is not involved in the project and
knows nothing about its design, corpus, or hypotheses. Before starting they
read only the Swales CARS annotation guidelines and the insight summary
described below. They are blind throughout to item provenance, to the
human-to-AI ratio, and to the pairing of the two versions of each paper.

\paragraph{Sampling and design.}
Sixty papers were sampled from the $76$ in the corpus, stratified by journal
quartile (Q1 $12$, Q2--Q4 $16$ each). Each sampled paper contributes two
items, its original introduction and one generated introduction for the same
paper, giving $120$ items in total. The generated introductions all come from Claude, which is
among the models whose structures come closest to the human originals
(\S\ref{sec:results:graph}). Two factors are crossed at the paper level, so
both of a paper's items share the same cell: the prompting setting of the
generated item (basic vs.\ CARS-guided) and whether the pair is shown with
automatic move annotations (labelled vs.\ unlabelled). Each of the four cells
holds $15$ papers, which yields $60$ items shown with
move annotations ($30$ original $+$ $15$ basic $+$ $15$ CARS-guided) and
$60$ shown as plain text. The $120$ items are presented in a single fixed
order in which a paper's two versions are always at least $20$ positions
apart, so that the two versions of one paper are never close enough for the
pairing to become visible.

\paragraph{Insight summary shown to the judge.}
Before starting, the judge reads the annotation guidelines and a short
summary of the corpus-level tendencies from \S\ref{sec:results:gen} and
\S\ref{sec:results:graph}. 

\begin{figure*}[t]
  \centering
  \includegraphics[width=\textwidth]{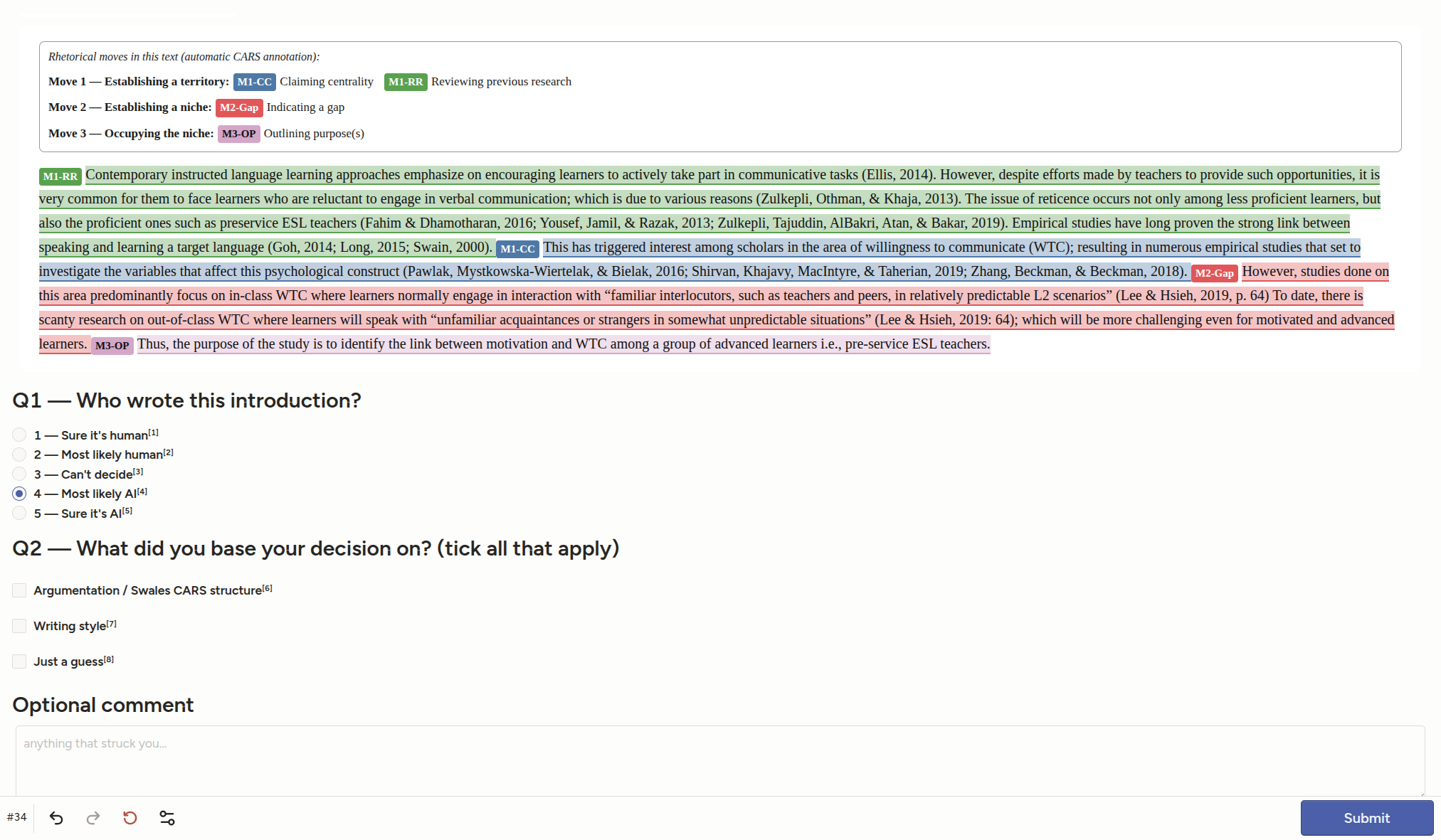}
  \caption{The annotation interface used in the human detection study
    (\S\ref{sec:analysis:detection}). A single introduction is shown, with its
    automatic CARS move annotations as coloured highlights in the labelled
    condition, above the five-point provenance question and the required
    decision-basis question. The colour key at the top lists only the moves
    that occur in the item on screen. The judge sees the whole introduction
    by scrolling; the text is cut off at the foot of the screen here.}
  \label{fig:detection-ui}
\end{figure*}

\paragraph{Response.}
Figure~\ref{fig:detection-ui} shows the interface. For each introduction the
judge answers a provenance question on a five-point scale (1 sure it is
human, 2 most likely human, 3 cannot decide, 4 most likely AI, 5 sure it is
AI) and a required question on the basis of the decision (the argumentation
or CARS structure, the writing style, or a guess; multiple may be selected).
In the labelled condition the automatic move annotations appear as coloured
highlights over the text, and the key above the introduction lists only the
moves that occur in the item on screen.

\paragraph{Analysis.}
The five-point ratings are read as a binary judgement by splitting the scale
at its midpoint, $1$ and $2$ counting as a vote for human and $4$ and $5$ as
a vote for AI. We report the number of correct judgements and the accuracy.
The reader never used the middle point of the scale.

\paragraph{Results.}
Table~\ref{tab:detection} reports the study. The reader was correct on $108$
of the $120$ items, and the two kinds of error are balanced: six generated
introductions passed as human and six originals were taken for machine
writing. 

\begin{table}[t]
\centering
\small
\setlength{\tabcolsep}{6pt}
\begin{tabular}{lcc}
\toprule
Condition & Items & Correct \\
\midrule
All items & 120 & 108 ($90\%$) \\
\midrule
\multicolumn{3}{l}{\textit{By move annotation (all items)}} \\
Labelled   & 60 & 54 ($90\%$) \\
Unlabelled & 60 & 54 ($90\%$) \\
\midrule
\multicolumn{3}{l}{\textit{Generated items, by prompting setting}} \\
Basic       & 30 & 24 ($80\%$) \\
CARS-guided & 30 & 30 ($100\%$) \\
\bottomrule
\end{tabular}
\caption{Results of the human detection study
(\S\ref{sec:analysis:detection}). \emph{Labelled}/\emph{Unlabelled}: whether the item
was shown with its automatic CARS move annotations as coloured highlights.
\emph{Basic}/\emph{CARS-guided}: the prompting setting of the generated
item. \emph{Correct} counts judgements that match the true provenance, so
the percentage is the accuracy.}
\label{tab:detection}
\end{table}

\paragraph{Prompting setting.}
All $30$ CARS-guided generations were identified as machine-written against
$24$ of the $30$ basic ones. The uniformity behind the effect is
visible in these $120$ items as well as in the corpus at large: $63\%$ of
the CARS-guided generations follow the plain $1 \rightarrow 2 \rightarrow 3$
move order against $13\%$ of the basic ones and $15\%$ of the originals, and
the $30$ CARS-guided introductions realise only $6$ distinct move structures
where the $60$ originals realise $29$.

\paragraph{Showing the move annotation made no difference.}
The reader was correct on $54$ of $60$ items in the labelled condition and
on $54$ of $60$ in the unlabelled one. 

We read this as a statement about what a reader needs rather than as
evidence that the annotations carry nothing. The reader named the structure
as the basis of the judgement on $58$ of the $60$ unlabelled items, and
reports having done the move analysis in their head, from the content alone,
on the items that came without annotations. The analysis was therefore being
done either way. What the reader had been given beforehand was the annotation
scheme and a short summary of how the two sources differ across the corpus,
and that turned out to be enough to act on.

\section{Prompts}
\label{sec:appendix:prompts}

This appendix reproduces all prompts used in the study: the system prompts
of the two generation settings (\S\ref{sec:data:gen}), the per-paper user
message they share, and the system prompt of the automatic annotator
(\S\ref{sec:method:evaluator}).

\subsection{Basic generation prompt}
\label{sec:appendix:genprompts}

The two generation settings share the same role framing, input description,
output rules, and JSON output shape. The basic setting uses the following
system prompt verbatim.

\begin{promptbox}{Basic Generation Prompt (system)}
You are an expert academic writer specialising in applied linguistics and
English language teaching (EFL/ESL). Your task is to write the
\textbf{Introduction} section of a research article.

You are given: \\
1.~The article's content --- its abstract and body sections (literature
review, methodology, results, discussion). This tells you what the study is
about, what it did, and what it found. IMPORTANT: the original Introduction
has been withheld on purpose; do not try to recall or reconstruct it ---
write a fresh one from the content you are given. \\
2.~The works the original authors cited in their introduction. Each entry
gives the authors, year and title, and includes an abstract where one was
available (some entries have only the title and authors --- judge from the
title what the work is about).

Rules: \\
-- Do NOT output a section heading --- only the introduction prose. \\
-- Cite sources in author--year format, e.g.\ ``Smith \& Jones, 2019''. Use
the works listed above where relevant; you may also cite other works where
the writing calls for it. \\
-- Length: write to the target length stated with the article below and stay
within the given range. Do not pad or over-elaborate to reach it, and do not
substantially exceed it.

Return ONLY a JSON object with this exact shape (no markdown, no
commentary): \{"introduction": "<the full introduction text>"\}
\end{promptbox}

\subsection{CARS-guided generation prompt}
\label{sec:appendix:carsprompt}

The CARS-guided setting extends the basic system prompt with the
move-planning instruction below and the CARS move and step definitions,
taken from the annotation manual \citep{sheikhasadi2025guidelines} and
inserted where \texttt{[CARS MOVE AND STEP DEFINITIONS]} appears.

\begin{promptbox}{CARS-Guided Generation Prompt (system, additions)}
You will write the introduction following \textbf{Swales' CARS (Create A
Research Space) model}, whose moves and steps are defined below.

Before writing, PLAN THE MOVES INTERNALLY: think through which CARS moves
and steps the introduction will use and in what sequence. The moves do NOT
have to appear in 1 $\rightarrow$ 2 $\rightarrow$ 3 order --- they recur and
interleave freely, and any ordering is acceptable (a move can be revisited,
e.g.\ returning to a Move 1 review after a Move 2 gap, and a real
introduction may open or close on any move). Keep this plan entirely to
yourself.

Then WRITE the introduction that realises your internal plan. Output ONLY
the introduction prose --- never the plan, and no move/step labels or
headings in the text.

\textbf{[CARS MOVE AND STEP DEFINITIONS]} \normalfont\small (the move and
step definitions of the annotation manual,
\citealp{sheikhasadi2025guidelines})
\end{promptbox}

\subsection{Generation user message}
\label{sec:appendix:usermsg}

In both generation settings, each paper is supplied in a user message
containing the article title, the article content, the cited works, and the
target length.

\begin{promptbox}{Per-Paper Generation User Message}
\# Article title \\
\{title\}

\# Article content (abstract + body; the original introduction is withheld) \\
\{body\}

\# Key prior works cited in the original introduction \\
\{references\}

\# Target length \\
\{length\_instruction\}

Now produce the Introduction. Return only the JSON object described in your
instructions.
\end{promptbox}

\subsection{Annotator prompt}
\label{sec:appendix:prompt}

The system prompt of the automatic annotator (\S\ref{sec:method:evaluator})
consists of the task framing (segmentation rules, the two-step labelling
procedure, the output format, and the numbered label set), the full
annotation manual \citep{sheikhasadi2025guidelines}, the operational
clarifications distilled from the error analyses of
\S\ref{sec:method:evaluator}, and the few-shot demonstrations. The box below
shows the structure of the prompt, with each detailed part summarised by a
bracketed placeholder. Each introduction is then supplied in a short user
message (``Please segment the following research article introduction into
coherent argumentative units. Return only the JSON --- no explanation.'').

\begin{promptbox}{Annotator Prompt (system)}
You are an expert in academic discourse analysis, specialising in Swales'
CARS (Create a Research Space) model for research article introductions.

\textbf{Task.} Following the annotation manual below: (1) split the
introduction into coherent argumentative segments; (2) for each segment,
first identify its MOVE, then its step; (3) give a one-sentence rationale
and output the class number.

\textbf{[SEGMENTATION INSTRUCTIONS]} \normalfont\small (five rules: cover
the whole introduction, never cut mid-sentence, segments verbatim and in
order, and a new segment whenever the communicative function
changes)\ttfamily

\textbf{[TWO-STEP LABELLING PROCEDURE]} \normalfont\small (decide the move
first --- Move 1/2/3 with their class-number ranges --- then the step
within that move)\ttfamily

\textbf{[OUTPUT FORMAT]} \normalfont\small (a JSON ``segments'' array whose
elements carry the verbatim ``text'', a one-sentence ``rationale'' written
before deciding the label, and an integer ``label'')\ttfamily

\textbf{[LABEL SET]} \normalfont\small (the numbered list of the eleven
steps, grouped by move)\ttfamily

\textbf{[ANNOTATION GUIDELINES]} \normalfont\small (the full annotation
manual of \citealp{sheikhasadi2025guidelines}, inserted verbatim)\ttfamily

\textbf{[OPERATIONAL CLARIFICATIONS]} \normalfont\small (tie-breaker rules
distilled from the error analyses of \S\ref{sec:method:evaluator}: the
decision order within Move 1, granularity and recurrence, one communicative
function per segment, continuing a tradition vs.\ indicating a gap,
frameworks and approaches, and purpose vs.\ present research)\ttfamily

\textbf{[WORKED EXAMPLES]} \normalfont\small (three gold-annotated
demonstration introductions --- coarse, median, and fine-grained --- rendered
in the output format, plus a constructed contrast paragraph for the three
Move-1 steps)\ttfamily

\textbf{Guiding principle.} Always classify segments by the communicative
function of each sentence --- what the author is doing rhetorically at that
point in the introduction --- not by surface cues, sentence boundaries, the
presence of connectors, or paragraph layout. A new segment is warranted only
when the communicative function genuinely changes.
\end{promptbox}

\end{document}